\documentclass[10pt,twocolumn,letterpaper]{article}

\usepackage{cvpr}
\usepackage{times}
\usepackage{epsfig}
\usepackage{graphicx}
\usepackage{amsmath}
\usepackage{amssymb}
\usepackage{array}
\usepackage{authblk}

\usepackage[pagebackref=true,breaklinks=true,letterpaper=true,colorlinks,bookmarks=false]{hyperref}

 \cvprfinalcopy 

\ifcvprfinal\fi
\everymath{\displaystyle}

\begin{document}
	
	\title{VommaNet: an End-to-End Network for Disparity Estimation from Reflective and Texture-less Light Field Images}
	
	\author[1,3]{Haoxin Ma}
\author[2]{Haotian Li}
\author[2]{Zhiwen Qian}
\author[1]{Shengxian Shi\thanks{kirinshi@sjtu.edu.cn}}
\author[3]{Tingting Mu}

	\affil[1]{School of Mechanical Engineering, Shanghai Jiao Tong University}
	\affil[2]{Vommatec Co. Ltd.}
	\affil[3]{School of Computer Science, The University of Manchester}
	
	\maketitle
	
	\begin{abstract}
		The precise combination of image sensor and micro-lens array enables lenslet light field cameras to record both angular and spatial information of incoming light, therefore, one can calculate disparity and depth from light field images. In turn, 3D models of the recorded objects can be recovered, which is a great advantage over other imaging system. However, reflective and texture-less areas in light field images have complicated conditions, making it hard to correctly calculate disparity with existing algorithms. To tackle this problem, we introduce a novel end-to-end network VommaNet to retrieve multi-scale features from reflective and texture-less regions for accurate disparity estimation.  Meanwhile, our network has achieved similar or better performance in other regions for both synthetic light field images and real-world data compared to the state-of-the-art algorithms. Currently, we achieve the best score for mean squared error (MSE) on HCI 4D Light Field Benchmark.
	\end{abstract}
	
	\section{Introduction}
	
	With recent developments in lenslet-based light field camera technology\cite{ng2005light}, especially those commercially available products from Lytro\cite{lytro} and Raytrix\cite{raytrix}, depth estimation from light field images has been a niche topic in computer vision. Based on the two-plane parameterization\cite{levoy2006light}, light field images can be used to generate multi-view images with slightly different view points and refocused images with different focal planes\cite{ng2006}. With these advantages, various algorithms\cite{jeon2015accurate, jeon2018depth, zhang2016robust} have been developed to estimate depth information from single light field image. Such depth information, when combined with sophisticated metric calibration techniques \cite{heinze2016automated, bok2017geometric}, could generate very dense point clouds as well as corresponding textures. This could be very attractive to 3D modeling and 3D geometry measurement community, especially for outdoor applications. 
	
	To further improve depth estimation accuracy for light field images, challenges induced by small viewing angle of lenslet-based light field camera need to be properly addressed.  A series of algorithms have therefore been proposed to solve the occlusions\cite{zhang2016robust, sheng2018occlusion, schilling2018trust}, narrow baseline\cite{jeon2015accurate}, and intrinsic component recovering\cite{alperovich2018light} difficulties. Although computationally expensive \cite{johannsen2017taxonomy}, these algorithms have been successfully applied in high-texture, non-reflective, and Lambertian surfaces. However, depth estimation from reflective and texture-less light field images remain a challenge for most of current algorithms. Attempts have been made to recover depth information for these regions with the help of shape-from-shading \cite{wu2011high, langguth2016shading, oxholm2014multiview}, but doing so would need prior knowledge of illumination (captured or estimated), and is generally limited to Lambertian surfaces or surfaces with uniform reflectance\cite{cui2017polarimetric}.  As pointed out by Zhu \etal\cite{zhu2017}, depth estimation from reflective and texture-less light field images has yet been fully studied and more attentions are needed before the light field imaging could become an attractive alternative for 3D modeling and 3D measurement community.
	
	In this paper, we propose a new end-to-end network that specifically addresses the problem of disparity estimation in reflective and texture-less area while maintains a similar or better performance in other regions compared to existing algorithms. For that purpose, our proposed network takes all of sub-aperture images (SAIs) directly as inputs to make full use of information recorded by light field cameras. Meanwhile, atrous convolutions of multiple scales are used to construct a feature pyramid in order to enlarge the receptive field of earlier layers to capture multi-scale features so that, for reflective and texture-less regions, the network can "take a step back" and view a larger picture. And in turn, the network can learn to infer disparity values for these regions from their neighborhood. Also, we use depthwise separable convolution and batch normalization to decrease parameter numbers, which can partially ease the computational burden.
	
	Our paper is organized as follows. In section 2, we introduce previous works in the fields of light field, depth estimation, and neural network; in section 3, we explain our network design in details; in section 4, we perform various experiments with both synthetic and real world light field images, and compare our results with those of some state-of-the-art algorithms; in section 5, we conclude our research and talk about future work and possible improvements.

\section{Related Work}

In this section, we will briefly introduce outstanding representatives of related approaches.

Light field can be re-sampled to SAIs with epipolar constraint, which can be calculated similar to stereo matching\cite{jeon2015accurate}. Therefore, also similar to stereo matching, depth from light field can be estimated based on correspondence. Jeon \etal\cite{jeon2015accurate} proposed a correspondence method based on phase shift theorem to solve narrow baseline, and improve the algorithm by using a cascade random forest to predict accurate depth value from matching costs\cite{jeon2018depth}. However, as Hane \etal\cite{hane2015direction} has demonstrated, correspondence based methods will not lead to a confident depth estimation in reflective and texture-less area, as many different disparities lead to low matching costs.

On the other hand, light field is commonly represented as multi-orientation epipolar plane images(EPIs)\cite{bolles1987epipolar}. Each of the lines on EPIs corresponds to the projection of a 3D point in space, and the various slopes can be represented as disparity, from which depth can be deducted. Based on the rich structure of EPIs, depth can be analyzed for more complex scenes, such as occlusion areas\cite{wanner2013reconstructing, johannsen2016sparse, wanner2013reconstructing}. Johannsen \etal\cite{johannsen2016sparse} used sparse coding on patches of the EPI to find those dictionary elements which best describe the patch. Zhang \etal\cite{zhang2016robust} proposed an EPI based Spinning Parallelogram Operator(SPO), which estimates the orientation of epipolar lines and is robust to occlusions. And Sheng \etal\cite{sheng2018occlusion} improved the method to achieve better accuracy by using multi-orientation EPIs. Schilling\cite{schilling2018trust} proposed a local optimization scheme based on the PatchMatch algorithm, which not only improved object boundaries, but also smooth surface reconstruction. 

\begin{figure}
	\begin{center}
		\includegraphics[width=1.0\linewidth]{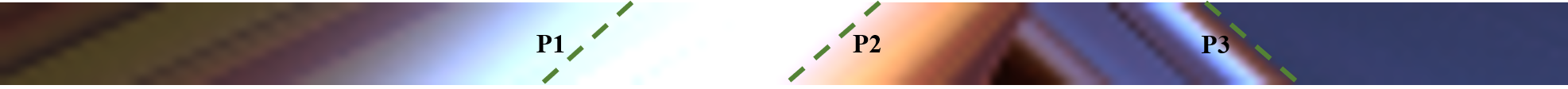}
	\end{center}
	\caption{EPI for reflective (between P1 and P2) and texture-less (left from P3) regions. It's clear that all pixels in these regions have the same RGB value.}
	\label{fig:epi}
\end{figure}

Furthermore, various recently proposed EPI based neural networks\cite{shin2018epinet, heber2016convolutional,heber2017neural,feng20173d, alperovich2018light} have shown promising performance in light field depth estimation. Heber \etal\cite{heber2016convolutional} used CNN to predict EPI line orientations, and then developed an end-to-end deep network architecture to predict depth\cite{heber2017neural}. Alperovich \etal\cite{alperovich2018light} present a fully convolutional autoencoder for light field images, which can be decoded in a variety of ways to acquire disparity map, diffuse, and specular intrinsic components. Feng \etal\cite{feng20173d} proposed FaceLFnet based on dense block and EPIs from horizontal and vertical SAIs. Shin and Jeon\cite{shin2018epinet} introduced a deep learning-based approach named EPINET for light field depth estimation that achieves accurate results and fast speed.  However, since EPI slopes are calculated primarily from neighboring pixel values, as demonstrated in Figure \ref{fig:epi}, EPI slopes cannot be correctly calculated for reflective and texture-less regions because all pixels in these areas have the same value.

The aforementioned algorithms are only feasible in ordinary non-reflective high-texture regions. For mirror-like reflective or low-texture surfaces, Wanner and Goldluecke\cite{wanner2014variational, wanner2013reconstructing} estimated the slope of epipolar lines by using the second order structure tensor to allow the reconstruction of multi-layered depth maps. They succeeded in accurately estimating depth for mirror-like surfaces and transparent objects. Tao \etal\cite{tao2017shape} combined the correspondence, defocus cue, and the shape of shading method to refine depth estimation results for Lambertian surfaces. Their method acquired accurate depth for surface of a model shell, a gloss and low-texture surface. Johannsen \etal\cite{johannsen2016sparse} proposed sparse light field coding to decompose the light field of specular surfaces into different superimposed layers, which can leverage the depth estimation for these regions.

As neural networks, especially convolutional neural network(CNN) demonstrate their advantages over traditional methods in numerous research fields, researchers in the field of neural network have carried out more and more works focusing on network structure and learning techniques.  Chen \etal\cite{chen2018deeplab} proposed atrous convolution which can enlarge the field-of-view, in other words, the receptive field of neural networks, without increasing the number of parameters or the amount of computation, and demonstrated its effectiveness in semantic segmentation. Also, depthwise separable convolution\cite{chollet2017xception,howard2017mobilenets} has been proposed to greatly decrease the parameter number while maintain a similar performance. And it has shown its feasibility in various fields such as image classification\cite{chollet2017xception}. Moreover, novel techniques of batch normalization\cite{ioffe2015batch} and residual neural networks\cite{he2016deep} have accelerated the training of deep neural networks while keeping them robust. Both methods have been proven effective by corresponding authors in the field of image classification. We are inspired by these advances and seek to take advantage of them to address the problem of accurate depth estimation for reflective and texture-less areas.

\begin{table}
	\begin{center}
		\begin{tabular}{| *4{>{\centering\arraybackslash}m{0.65in}|} @{}m{0pt}@{}}
			\hline
			VommaNet\_&9&25&81\\
			\hline\hline
			Input views & 
			\includegraphics[width=\linewidth]{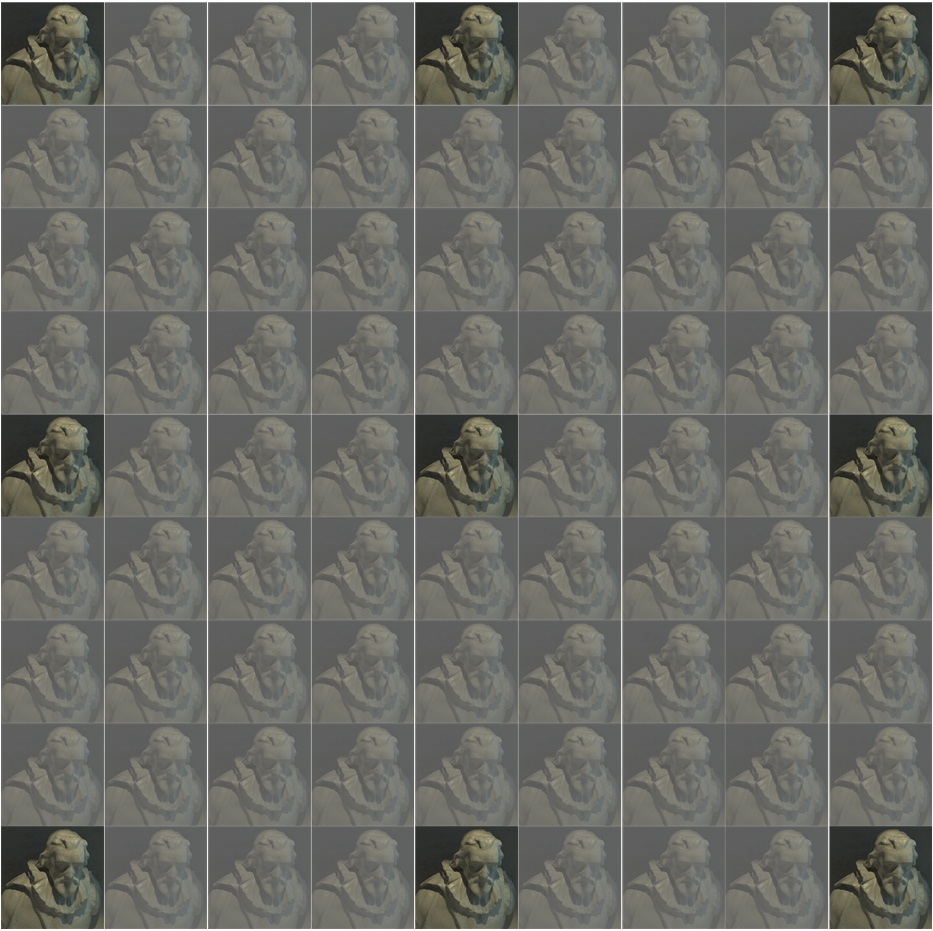}
			&\includegraphics[width=\linewidth]{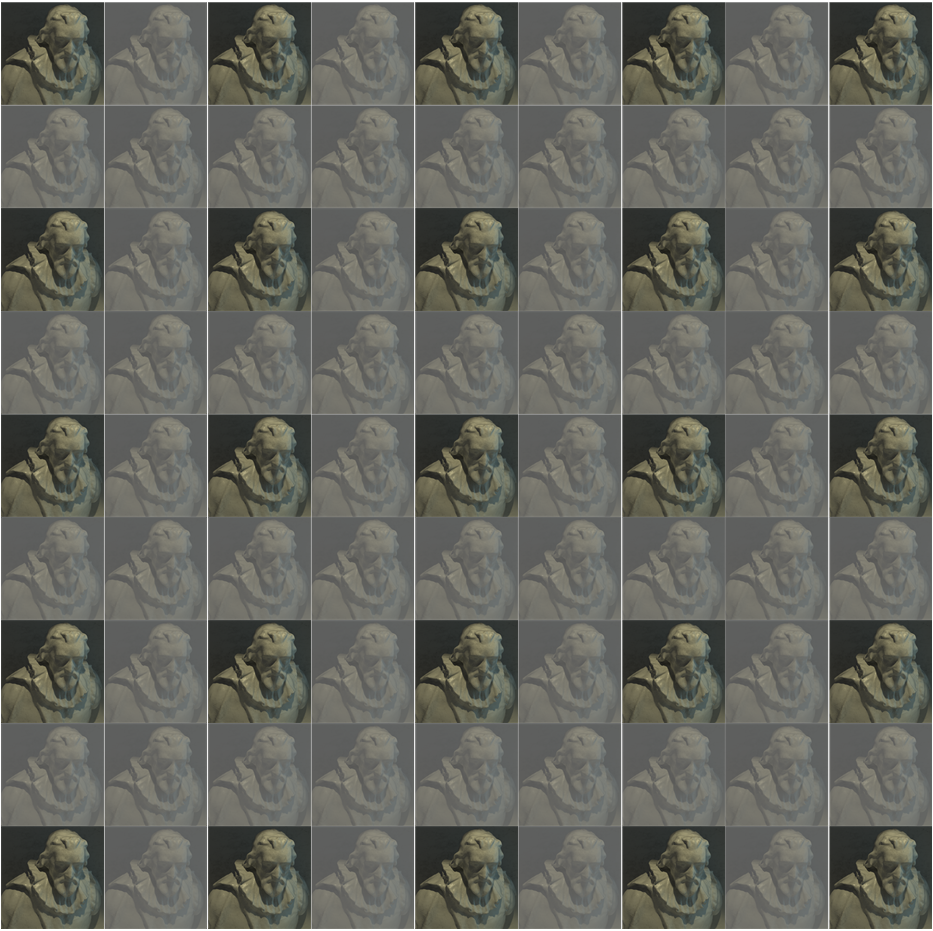}
			&\includegraphics[width=\linewidth]{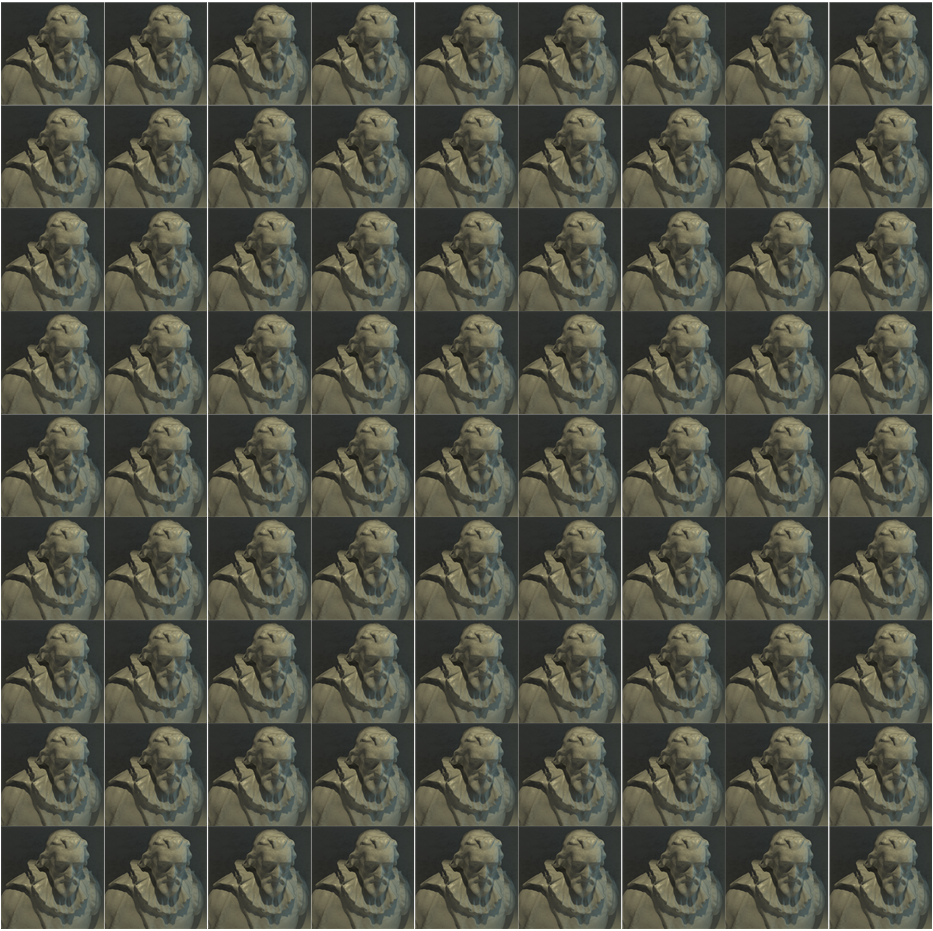}
			\\
			\hline
			SAI No. & 3$\times$3 & 5$\times$5 & 9$\times$9\\
			\hline
			Param No./Million & 0.83 & 0.99 & 1.57\\
			\hline
			MSE $\times 10^-2$ & 1.53 & 1.16 & 0.82\\
			\hline
			Runtime/s & 0.834 & 1.679 & 2.043\\
			\hline
		\end{tabular}
	\end{center}
	\caption{Performance comparison between different input patterns. MSE and runtime are computed on validation data of 512$\times$512 in size.}
	\label{table:inputs}
\end{table}

\section{Our Method}

As demonstrated above, depth for reflective and texture-less areas cannot be accurately estimated if we only examine these areas locally. However, if we take a step back and examine a larger region, we will find that, the disparities for the edges of these areas can be easily calculated. Also, since the disparity values inside an object should be continuous, we can let the network estimate the disparities of reflective and texture-less areas based on those of their edges. Therefore, if we can enable the network to "step back" and "see" a larger region and combine this information with local features extracted from smaller regions, there should be performance improvement for texture-less and reflective areas.

In other words, our network should be able to extract multi-scale features from light field images. However, this means we need to perform multi-scale convolutions, which may lead to heavy computational burden. Therefore, we also need to decrease parameter number when designing our network.

\subsection{Input}

In order to let the network extract features from both small and large regions, we concatenate all SAIs in channel axis and take it in directly as input and apply convolution layers with both small and large receptive field. This way, the network will be able to extract both local and global features directly from SAIs.


\subsection{Network Design}

\begin{figure*}
	\begin{center}
		\includegraphics[width=0.8\linewidth]{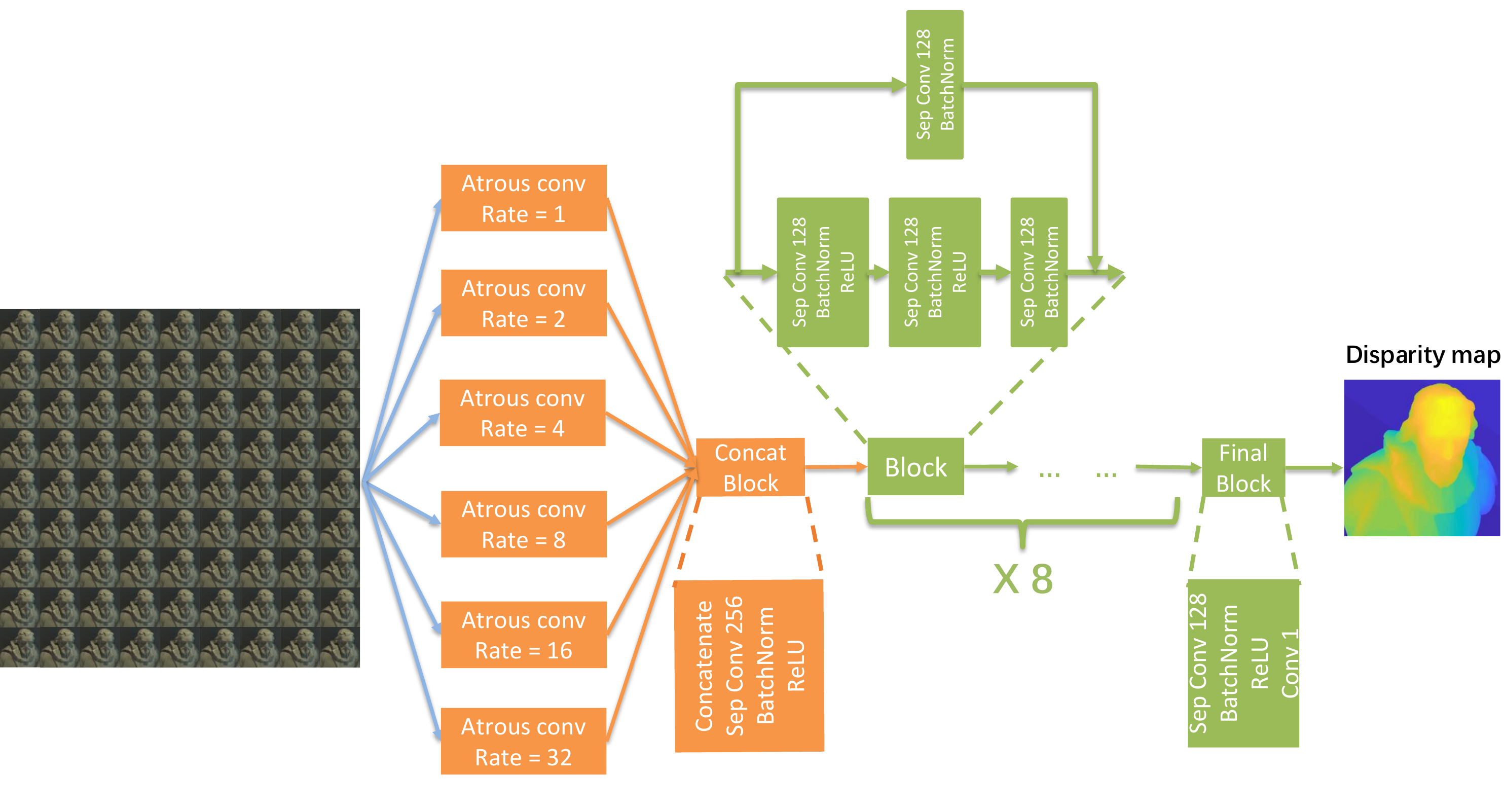}
	\end{center}
	\caption{The structure of our network.}
	\label{fig:structure}
\end{figure*}

As stated previously, we aim to enlarge the receptive fields of the network in earlier layers, we design our network as shown in Figure \ref{fig:structure}.

First, we use a feature pyramid consisted of atrous convolution with increasing dilation rates to extract multi-scale features so that our network can infer disparity values for reflective and texture-less areas from their surroundings. The feature pyramid we proposed consists of 6 atrous convolution layers with dilation rates of 1(no dilation), 2, 4, 8, and 16 separately. We adopt this structure because, as demonstrated by Chen \etal\cite{chen2018deeplab}, atrous convolution up-samples the convolutional kernels by padding zeros in between trainable parameters, which can effectively enlarge receptive field while keeping a rather low parameter number and computation amount.

The outputs from different layers of the pyramid, in other words low level multi-scale features, are concatenated along the channel axis and passed to a depthwise separable convolution layer to encode these outputs into higher-level features. After this, we apply a series of residual blocks followed by one single convolution layer to have the final output. One residual block has two passes, one shortcut of a depthwise separable convolution layer, and another pass consisted of three consecutive depthwise separable convolution layers. The last convolution layer has one single 1$\times$1 kernel while all other convolutional kernels are of 3$\times$3 in size. The outputs from two passes are added together to get the output of this one residual block.

We choose to use residual blocks mainly for two reasons. First, our network consists of a large number of layers, which makes it prone to the vanishing gradient problem, and as demonstrated by He \etal\cite{he2016deep}, residual structure can avoid this problem by re-introducing outputs from shallower layers in the network to compensate for the vanishing data. Second, deeper network means larger number of parameters, which increase computational burden. For that reason, we use depthwise separable convolutions in substitution of normal convolutions to decrease parameter number and speed up training.

All convolution layers in our network is followed by a batch normalization layer and a ReLU activation layer except for the last one.


For comparison and evaluation purpose, we train three different networks, which take in as input all, 25, and 9 out of all 81 SAIs separately. We perform experiments on synthetic data to compare the performance of networks with these different input patterns. This is demonstrated in Table \ref{table:inputs}. These three networks are noted as VommaNet\_81, VommaNet\_25, and VommaNet\_9 hereafter. Among the three networks we have trained, there is noticeable increment in performance as input SAI number increases. Also, the parameter number, training time, and runtime of the network rises significantly.

\subsection{Loss Function}

Since there are shortcuts in the residual blocks, texture of the input images may be preserved in the final output. Therefore, an effective loss function should enforce not only accuracy but structural similarity between network output and ground truth as well.

Most of the previous studies employ mean absolute error(MAE) between network estimation $d_{i}$ and its ground truth $g_{i}$ as loss function to enforce accuracy for network output:

$$l_{MAE} = \sum_{i=1}^{N} \frac{|D_{i}|}{N}$$

where $N$ is the total number of pixels, and $D_{i}=d_{i}-g_{i}$ is the difference between network estimation and its ground truth at the $i^{th}$ pixel. However, as illustrated in \cite{hu2018revisiting}, this loss is insensitive to distortion and blur of edges. Therefore, we employ the following loss to penalize errors around edges more:

$$l_{grad} = \sum_{i=1}^{N} \frac{|\nabla_{x}(D_{i})|+|\nabla_{y}(D_{i})|}{N}$$

where $\nabla_{x}$ is spatial gradient in $x$-axis, and  $\nabla_{y}$ is that in $y$-axis. To further improve fine details of depth maps, we consider yet another loss from \cite{hu2018revisiting}, which measures accuracy of the normal to the surface of an estimated depth map with respect to its ground truth:

$$l_{normal} = 1 - \sum_{i=1}^{N} \frac{\cos <\overrightarrow{n}^{d}_{i}, \overrightarrow{n}^{g}_{i}>}{N}$$

where $\overrightarrow{n}^{d}_{i} = (-\nabla_{y}d_{i}, -\nabla_{x}d_{i}, 1)$, and $\overrightarrow{n}^{g}_{i} = (-\nabla_{y}g_{i}, -\nabla_{x}g_{i}, 1)$. Finally, we use weighted sum of the above loss functions to train our network.

$loss = \lambda_{1}l_{MAE}+\lambda_{2}l_{grad}+\lambda_{3}l_{normal} $

where $\lambda_{1}, \lambda_{2}, \lambda_{3}$ are coefficients for different terms.

\subsection{Training Details}

We use data provided by \cite{alperovich2018light} along with the additional data provided by the benchmark \cite{honauer2016dataset} as training data. Since the data amount is not very large, we augment the data by flipping, color inversion, and cropping into mini-batch. To generate training data, we flip the data up-down, left-right, and up-down plus left-right, then invert image color, and finally cut them into mini-batches of size 64.

The learning rate begins from 0.001 and decays every 10 epochs by a factor of 0.5 until it falls below $1\times10^{-7}$. The training process takes about three days with Intel E5-2603 v4 @1.7GHz, 64GB RAM, and Nvidia GeForce GTX 1080Ti. And for loss function, we set all coefficients to 1.

\section{Experiments}

We perform a number of experiments on both synthetic data and real world light field images captured by a Lytro Illum camera, where we compare results from \cite{jeon2015accurate}, \cite{zhang2016robust}, and \cite{shin2018epinet} with those from our network. Note that for the existing algorithms we comparing with, we directly use the codes and evaluation metrics scores that the authors published.

\subsection{Synthetic Data}

We compared our algorithm with other state-of-the-art algorithms among the benchmark data provided by \cite{honauer2016dataset}, and the mean squared error values and runtime(reported by corresponding authors) are listed in Table \ref{table:results}.

\begin{figure*}
	\begin{center}
		\includegraphics[width=0.8\linewidth]{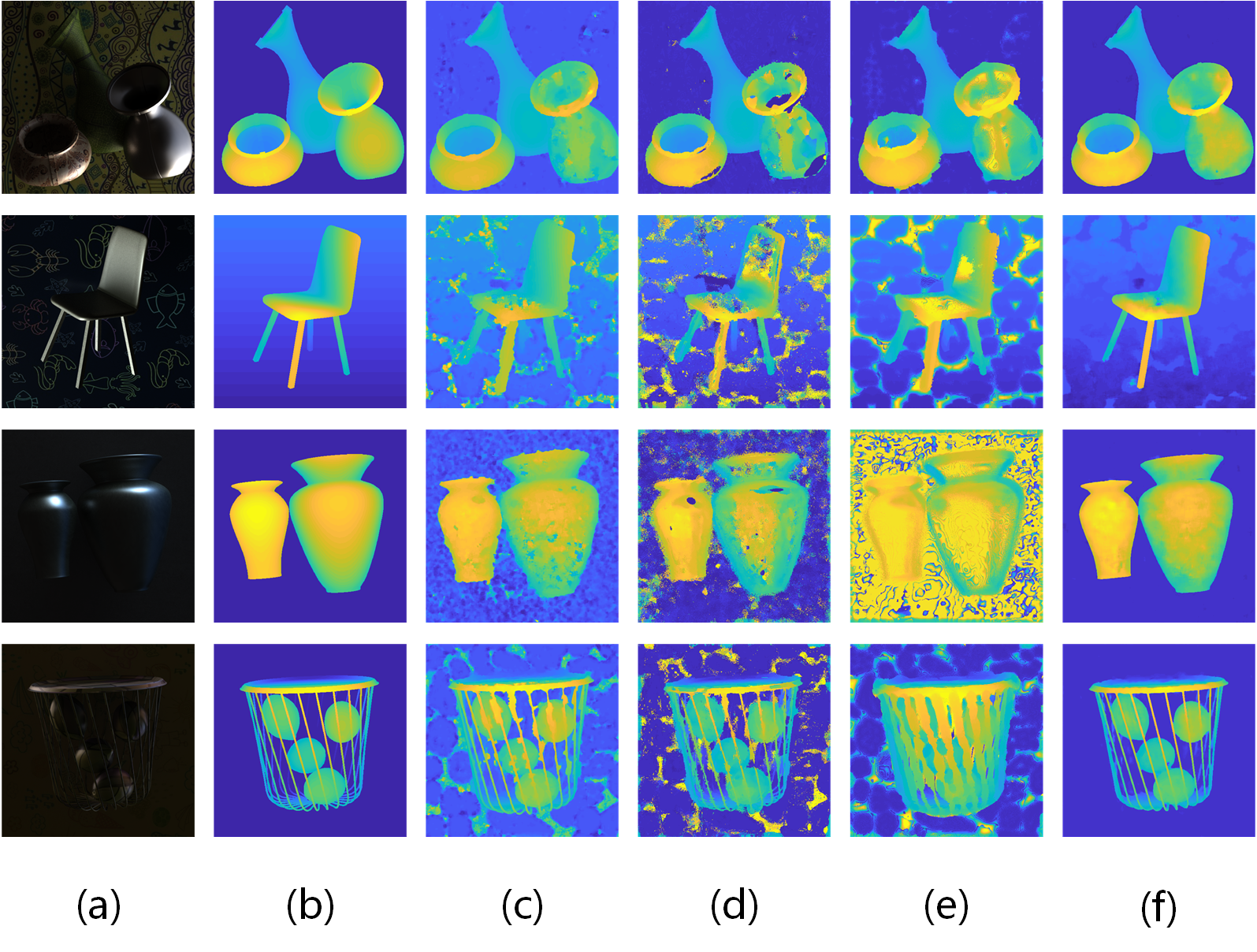}
	\end{center}
	\caption{Synthetic data results. (a)central view (b)ground truth (c)\cite{shin2018epinet} (d)\cite{zhang2016robust} (e)\cite{jeon2015accurate} (f)Ours(VommaNet\_81)}
	\label{fig:synthetic}
\end{figure*}

Meanwhile, for qualitative comparison, we run these algorithms on four scenes selected from \cite{alperovich2018light}, and the results are presented in Figure \ref{fig:synthetic}. Note that for our network, these scenes are excluded from our training data. As can be seen from the results, existing algorithms perform badly at reflective and texture-less areas such as the surfaces of bottles and chair, while our network yield generally good results.

\begin{table}
	\begin{center}
		\begin{tabular}{|l|c|c|}
			\hline
			Method & MSE $\times10^{-2}$ & Runtime/s \\
			\hline\hline
			\cite{zhang2016robust} & 3.968 & 2115.407\\
			\cite{shin2018epinet} & 2.521 & 2.041\\
			Ours(VommaNet\_81) & 2.218 & 2.043\\
			\hline
		\end{tabular}
	\end{center}
	\caption{Results comparison. Runtime is reported by author. For both scores, lower is better.}
	\label{table:results}
\end{table}

\subsection{Real World Data}

\begin{figure*}
	\begin{center}
		\includegraphics[width=0.8\linewidth]{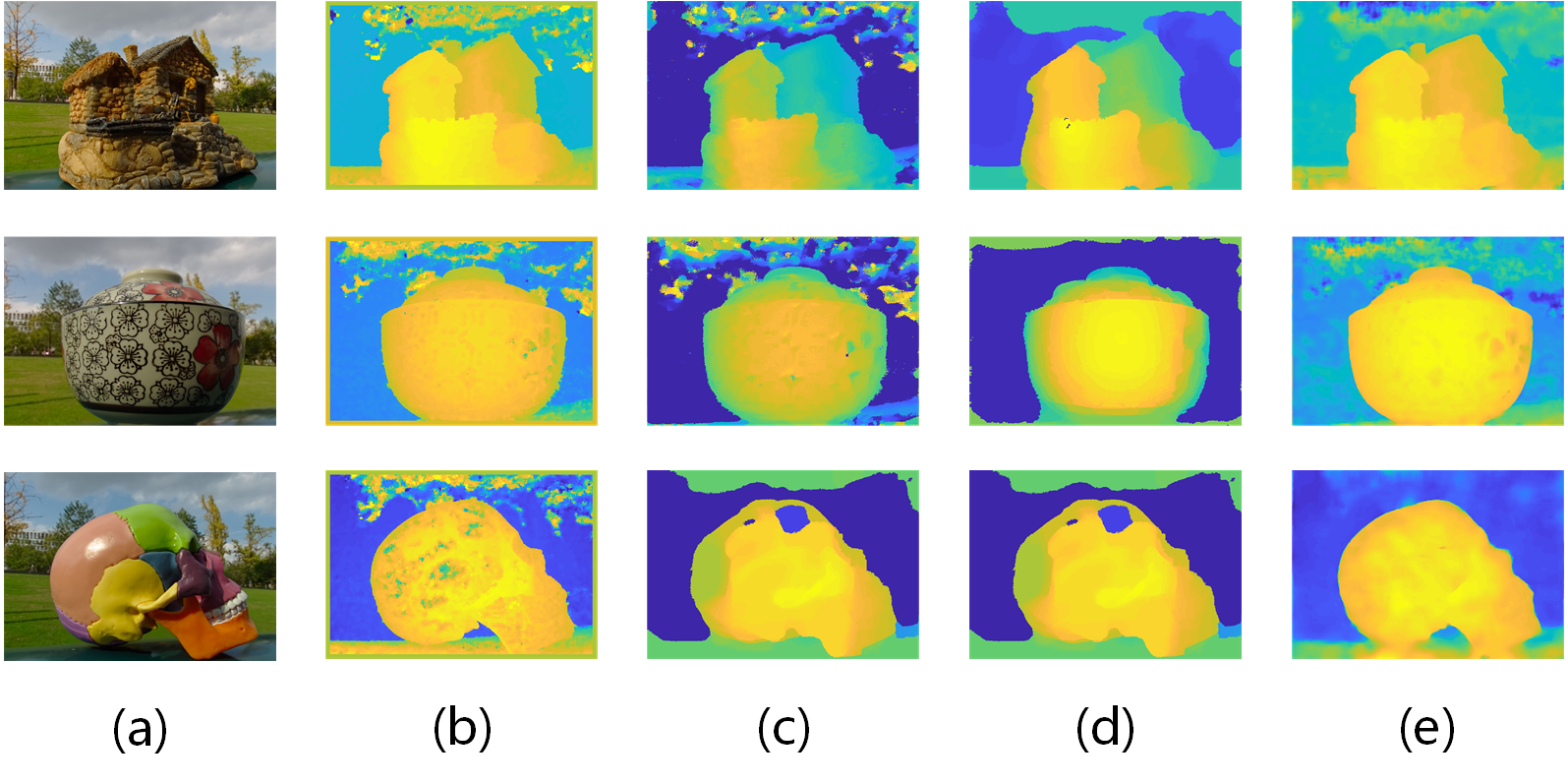}
	\end{center}
	\caption{Real world results. (a)thumbnail (b)\cite{shin2018epinet} (c)\cite{zhang2016robust} (d)\cite{jeon2015accurate} (e)Ours(VommaNet\_81)}
	\label{fig:lytro}
\end{figure*}

We also carry out experiments with hand held Lytro Illum cameras. We capture three different real world scenes where texture-less and reflective areas as well as ordinary ones are present, and process them with different state-of-the-art algorithms as well as our proposed networks. The results are shown in Figure \ref{fig:lytro}.

From the first two rows of Figure \ref{fig:lytro}, we can see that our network yields better results in ordinary scenes, especially at edges where other algorithms tend to be spiky while ours sharp and clear. Also, from the third row we can see that for reflective and texture-less areas, \eg the rear of the skull, result from our network remains smooth and accurate while existing algorithms clearly fail, generating absurd values. And the same as that with synthetic data, there is noticeable performance gain as input SAI number increases among the three networks we trained.

\section{Conclusion}

As presented previously, our algorithm has good performance in reflective and texture-less areas as well as ordinary ones. Meanwhile, our network has achieved better overall accuracy than existing methods while maintaining similar runtime.

Although our network performs well in reflective and texture-less areas, it does bad in preserve details. Complicated structures within objects may be blurry in our network, while texture from background or on object surfaces may be preserved to some extent, therefore, our network yields pretty large score for bad pixel metrics. Also, we trained different models with different input image numbers, and this can be improved by modify the network to be recurrent, similar to \cite{choy20163d}. This way, we will be able to train one single model for different numbers of input images, further extending the application of our network.

{\small
\bibliographystyle{ieee}
\bibliography{egbib}

\begin{thebibliography}{10}\itemsep=-1pt

\bibitem{lytro}
{Lytro.} illum.
\newblock \url{https://illum.lytro.com/illum}.
\newblock Accessed: 2018-11-01.

\bibitem{raytrix}
{Raytrix.} 3d light field camera technology.
\newblock \url{http://www.raytrix.de/}.
\newblock Accessed: 2018-11-01.

\bibitem{alperovich2018light}
A.~Alperovich, O.~Johannsen, M.~Strecke, and B.~Goldluecke.
\newblock Light field intrinsics with a deep encoder-decoder network.
\newblock {\em representations}, 28:15, 2018.

\bibitem{bok2017geometric}
Y.~Bok, H.-G. Jeon, and I.~S. Kweon.
\newblock Geometric calibration of micro-lens-based light field cameras using
  line features.
\newblock {\em IEEE transactions on pattern analysis and machine intelligence},
  39(2):287--300, 2017.

\bibitem{bolles1987epipolar}
R.~C. Bolles, H.~H. Baker, and D.~H. Marimont.
\newblock Epipolar-plane image analysis: An approach to determining structure
  from motion.
\newblock {\em International journal of computer vision}, 1(1):7--55, 1987.

\bibitem{chen2018deeplab}
L.-C. Chen, G.~Papandreou, I.~Kokkinos, K.~Murphy, and A.~L. Yuille.
\newblock Deeplab: Semantic image segmentation with deep convolutional nets,
  atrous convolution, and fully connected crfs.
\newblock {\em IEEE transactions on pattern analysis and machine intelligence},
  40(4):834--848, 2018.

\bibitem{chollet2017xception}
F.~Chollet.
\newblock Xception: Deep learning with depthwise separable convolutions.
\newblock {\em arXiv preprint}, pages 1610--02357, 2017.

\bibitem{choy20163d}
C.~B. Choy, D.~Xu, J.~Gwak, K.~Chen, and S.~Savarese.
\newblock 3d-r2n2: A unified approach for single and multi-view 3d object
  reconstruction.
\newblock In {\em European conference on computer vision}, pages 628--644.
  Springer, 2016.

\bibitem{cui2017polarimetric}
Z.~Cui, J.~Gu, B.~Shi, P.~Tan, and J.~Kautz.
\newblock Polarimetric multi-view stereo.
\newblock In {\em Proc. of Computer Vision and Pattern Recognition (CVPR)},
  2017.

\bibitem{feng20173d}
M.~Feng, S.~Z. Gilani, Y.~Wang, and A.~Mian.
\newblock 3d face reconstruction from light field images: A model-free
  approach.
\newblock {\em arXiv preprint arXiv:1711.05953}, 2017.

\bibitem{hane2015direction}
C.~Hane, L.~Ladicky, and M.~Pollefeys.
\newblock Direction matters: Depth estimation with a surface normal classifier.
\newblock In {\em Proceedings of the IEEE Conference on Computer Vision and
  Pattern Recognition}, pages 381--389, 2015.

\bibitem{he2016deep}
K.~He, X.~Zhang, S.~Ren, and J.~Sun.
\newblock Deep residual learning for image recognition.
\newblock In {\em Proceedings of the IEEE conference on computer vision and
  pattern recognition}, pages 770--778, 2016.

\bibitem{heber2016convolutional}
S.~Heber and T.~Pock.
\newblock Convolutional networks for shape from light field.
\newblock In {\em Proceedings of the IEEE Conference on Computer Vision and
  Pattern Recognition}, pages 3746--3754, 2016.

\bibitem{heber2017neural}
S.~Heber, W.~Yu, and T.~Pock.
\newblock Neural epi-volume networks for shape from light field.
\newblock In {\em Proceedings of International Conference on Computer Vision
  (ICCV)}, volume~2, 2017.

\bibitem{heinze2016automated}
C.~Heinze, S.~Spyropoulos, S.~Hussmann, and C.~Perwass.
\newblock Automated robust metric calibration algorithm for multifocus
  plenoptic cameras.
\newblock {\em IEEE Transactions on Instrumentation and Measurement},
  65(5):1197--1205, 2016.

\bibitem{honauer2016dataset}
K.~Honauer, O.~Johannsen, D.~Kondermann, and B.~Goldluecke.
\newblock A dataset and evaluation methodology for depth estimation on 4d light
  fields.
\newblock In {\em Asian Conference on Computer Vision}, pages 19--34. Springer,
  2016.

\bibitem{howard2017mobilenets}
A.~G. Howard, M.~Zhu, B.~Chen, D.~Kalenichenko, W.~Wang, T.~Weyand,
  M.~Andreetto, and H.~Adam.
\newblock Mobilenets: Efficient convolutional neural networks for mobile vision
  applications.
\newblock {\em arXiv preprint arXiv:1704.04861}, 2017.

\bibitem{hu2018revisiting}
J.~Hu, M.~Ozay, Y.~Zhang, and T.~Okatani.
\newblock Revisiting single image depth estimation: Toward higher resolution
  maps with accurate object boundaries.
\newblock {\em arXiv preprint arXiv:1803.08673}, 2018.

\bibitem{ioffe2015batch}
S.~Ioffe and C.~Szegedy.
\newblock Batch normalization: Accelerating deep network training by reducing
  internal covariate shift.
\newblock {\em arXiv preprint arXiv:1502.03167}, 2015.

\bibitem{jeon2018depth}
H.-G. Jeon, J.~Park, G.~Choe, J.~Park, Y.~Bok, Y.~W. Tai, and I.~S. Kweon.
\newblock Depth from a light field image with learning-based matching costs.
\newblock {\em IEEE Transactions on Pattern Analysis and Machine Intelligence},
  2018.

\bibitem{jeon2015accurate}
H.-G. Jeon, J.~Park, G.~Choe, J.~Park, Y.~Bok, Y.-W. Tai, and I.~So~Kweon.
\newblock Accurate depth map estimation from a lenslet light field camera.
\newblock In {\em Proceedings of the IEEE conference on computer vision and
  pattern recognition}, pages 1547--1555, 2015.

\bibitem{johannsen2017taxonomy}
O.~Johannsen, K.~Honauer, B.~Goldluecke, A.~Alperovich, F.~Battisti, Y.~Bok,
  M.~Brizzi, M.~Carli, G.~Choe, M.~Diebold, et~al.
\newblock A taxonomy and evaluation of dense light field depth estimation
  algorithms.
\newblock In {\em CVPR Workshops}, pages 1795--1812, 2017.

\bibitem{johannsen2016sparse}
O.~Johannsen, A.~Sulc, and B.~Goldluecke.
\newblock What sparse light field coding reveals about scene structure.
\newblock In {\em Proceedings of the IEEE Conference on Computer Vision and
  Pattern Recognition}, pages 3262--3270, 2016.

\bibitem{langguth2016shading}
F.~Langguth, K.~Sunkavalli, S.~Hadap, and M.~Goesele.
\newblock Shading-aware multi-view stereo.
\newblock In {\em European Conference on Computer Vision}, pages 469--485.
  Springer, 2016.

\bibitem{levoy2006light}
M.~Levoy.
\newblock Light fields and computational imaging.
\newblock {\em Computer}, 39(8):46--55, 2006.

\bibitem{ng2006}
R.~Ng et~al.
\newblock {\em Digital light field photography}.
\newblock stanford university Stanford, CA, 2006.

\bibitem{ng2005light}
R.~Ng, M.~Levoy, M.~Br{\'e}dif, G.~Duval, M.~Horowitz, P.~Hanrahan, et~al.
\newblock Light field photography with a hand-held plenoptic camera.
\newblock {\em Computer Science Technical Report CSTR}, 2(11):1--11, 2005.

\bibitem{oxholm2014multiview}
G.~Oxholm and K.~Nishino.
\newblock Multiview shape and reflectance from natural illumination.
\newblock In {\em Proceedings of the IEEE Conference on Computer Vision and
  Pattern Recognition}, pages 2155--2162, 2014.

\bibitem{schilling2018trust}
H.~Schilling, M.~Diebold, C.~Rother, and B.~J{\"a}hne.
\newblock Trust your model: Light field depth estimation with inline occlusion
  handling.
\newblock In {\em Proceedings of the IEEE Conference on Computer Vision and
  Pattern Recognition}, pages 4530--4538, 2018.

\bibitem{sheng2018occlusion}
H.~Sheng, P.~Zhao, S.~Zhang, J.~Zhang, and D.~Yang.
\newblock Occlusion-aware depth estimation for light field using
  multi-orientation epis.
\newblock {\em Pattern Recognition}, 74:587--599, 2018.

\bibitem{shin2018epinet}
C.~Shin, H.-G. Jeon, Y.~Yoon, I.~S. Kweon, and S.~J. Kim.
\newblock Epinet: A fully-convolutional neural network using epipolar geometry
  for depth from light field images.
\newblock In {\em Proceedings of the IEEE Conference on Computer Vision and
  Pattern Recognition}, pages 4748--4757, 2018.

\bibitem{tao2017shape}
M.~W. Tao, P.~P. Srinivasan, S.~Hadap, S.~Rusinkiewicz, J.~Malik, and
  R.~Ramamoorthi.
\newblock Shape estimation from shading, defocus, and correspondence using
  light-field angular coherence.
\newblock {\em IEEE transactions on pattern analysis and machine intelligence},
  39(3):546--560, 2017.

\bibitem{wanner2013reconstructing}
S.~Wanner and B.~Goldluecke.
\newblock Reconstructing reflective and transparent surfaces from epipolar
  plane images.
\newblock In {\em German Conference on Pattern Recognition}, pages 1--10.
  Springer, 2013.

\bibitem{wanner2014variational}
S.~Wanner and B.~Goldluecke.
\newblock Variational light field analysis for disparity estimation and
  super-resolution.
\newblock {\em IEEE transactions on pattern analysis and machine intelligence},
  36(3):606--619, 2014.

\bibitem{wu2011high}
C.~Wu, B.~Wilburn, Y.~Matsushita, and C.~Theobalt.
\newblock High-quality shape from multi-view stereo and shading under general
  illumination.
\newblock 2011.

\bibitem{zhang2016robust}
S.~Zhang, H.~Sheng, C.~Li, J.~Zhang, and Z.~Xiong.
\newblock Robust depth estimation for light field via spinning parallelogram
  operator.
\newblock {\em Computer Vision and Image Understanding}, 145:148--159, 2016.

\bibitem{zhu2017}
H.~Zhu, Q.~Wang, and J.~Yu.
\newblock Light field imaging: models, calibrations, reconstructions, and
  applications.
\newblock {\em Frontiers of Information Technology {\&} Electronic
  Engineering}, 18(9):1236--1249, Sep 2017.

\end{thebibliography}
}

\end{document}